\documentclass[sn-mathphys,Numbered]{sn-jnl}% Math and Physical Sciences Reference Style
\usepackage{graphicx}%
\usepackage{multirow}%
\usepackage{amsmath,amssymb,amsfonts}%
\usepackage{amsthm}%
\usepackage{mathrsfs}%
\usepackage[title]{appendix}%
\usepackage{xcolor}%
\usepackage{textcomp}%
\usepackage{manyfoot}%
\usepackage{booktabs}%
\usepackage{algorithm}%
\usepackage{algorithmicx}%
\usepackage{algpseudocode}%
\usepackage{listings}%
\usepackage{parskip}
\usepackage{hyperref}

\theoremstyle{thmstyleone}%
\theoremstyle{thmstyletwo}%

\theoremstyle{thmstylethree}%

\begin{document}

\title[Article Title]{Minimum Levels of Interpretability for Artificial Moral Agents}


\author*[1,2,3]{\fnm{Avish} \sur{Vijayaraghavan}}\email{av1017@ic.ac.uk}

\author[3]{\fnm{Cosmin} \sur{Badea}}%\email{iiauthor@gmail.com}

\affil[1]{\orgdiv{Section of Bioinformatics, Division of Systems Medicine, Department of Metabolism, Digestion and Reproduction}, \orgname{Imperial College London}, \orgaddress{\city{London}, \country{UK}}}

\affil[2]{\orgdiv{UKRI Centre for Doctoral Training in AI for Healthcare}, \orgname{Imperial College London}, \orgaddress{\city{London}, \country{UK}}}

\affil[3]{\orgdiv{Department of Computing}, \orgname{Imperial College London}, \orgaddress{\city{London}, \country{UK}}}


\abstract{ As artificial intelligence (AI) models continue to scale up, they are becoming more capable and integrated into various forms of decision-making systems. For models involved in moral decision-making, also known as artificial moral agents (AMA), interpretability provides a way to trust and understand the agent's internal reasoning mechanisms for effective use and error correction. In this paper, we provide an overview of this rapidly-evolving sub-field of AI interpretability, introduce the concept of the Minimum Level of Interpretability (MLI) and recommend an MLI for various types of agents, to aid their safe deployment in real-world settings.

\textbf{Abbreviations.} AI = artificial intelligence, AMA = artificial moral agent, BU = bottom-up, GPT = generative pre-trained transformer, IML = interpretable machine learning (or interpretability), LLM = large language model, MDM = moral decision-making, ML = machine learning, MLI = minimum level of interpretability, TD = top-down.
}

\maketitle

% \large

\section{Introduction}\label{sec1}

The deployment of consumer-facing generative artificial intelligence (AI) models such as Midjourney and ChatGPT has raised important questions on the ethics \cite{sallam2023chatgpt} and consequences of widespread access to AI technologies \cite{eloundou2023gpts}. Tracing the evolution of these models over the past five years \cite{yang2023harnessing}, it is likely that we will soon see multi-modal general-purpose models \cite{reed2022generalist, ibarz2022generalist, jablonka2022gpt, wang2022medclip, acosta2022multimodal} available to the public. As these models begin operating with higher autonomy and become integrated into existing applications \cite{mostafa2019adjustable, cervantes2016autonomous, mialon2023augmented} (e.g. ChatGPT with plugins, AI vision models within self-driving cars), they will play a greater role in many aspects of human decision-making \cite{nashed2023fairness, cervantes2020artificial}. A fundamental subset of human decision-making is moral decision-making (MDM). MDM comes in many forms - some examples include predicting whether criminals will reoffend \cite{chouldechova2017fair}, deciding appropriate treatment plans for patients \cite{fatemi2021medical}, and executing military defense strategies in a way that is compliant with original mission orders \cite{brutzman2018ethical}. MDM is difficult because it often involves weighing competing values in complex and ambiguous situations \cite{haidt2008morality}. Where other types of decision-making may be based in pragmatic considerations, like efficiency or performance, MDM requires making judgements about what is right and wrong.

AI models that are involved in MDM are called artificial moral agents (AMAs). For these types of decisions, it is imperative that we have high levels of agent understanding so that errors can be corrected swiftly and performance better aligned with human values to prevent unintended and potentially harmful agent effects \cite{mostafa2019adjustable, cervantes2020artificial}. ``Understanding'' an agent can take on different levels of complexity \cite{lipton2018mythos, watson2022conceptual} which require different AMA constructions \cite{cervantes2020artificial} for effective deployment. Here, effective deployment means finding a model of appropriate capacity for its task so it can be deployed and updated as efficiently as possible in the real-world. This understanding of agent behaviour is enabled through the field of interpretable machine learning (IML, a.k.a. interpretability) which helps make AI models more trustworthy and transparent \cite{lipton2018mythos, rudin2019stop}.

Taking the necessity of interpretability in AMAs as a spectrum, we frame our discussion less around whether AMAs need interpretability (i.e. a binary decision), and more around the Minimum Level of Interpretability (MLI) for different AMA constructions. As such, this paper revolves around three key concepts: \textbf{different AMA constructions}, \textbf{different levels of interpretability}, and \textbf{how AMA capability is altered by different levels of interpretability}. Much of machine ethics research concerns aligning ethical schemas with machines \cite{martinho2021perspectives}. Our work continues down this line, aiming to bridge technical aspects of interpretability with AMA construction for smoother deployment. An important disclaimer is that this field is nascent and we are proposing general safety rules based on limited evidence - the MLI will evolve as more evidence is gathered for different AMA use cases. Our scope is limited to a computational understanding of moral decision-making in AMAs and we do not consider the problem of where responsibility falls for agent decisions \cite{hammond2021learning}. 

\section{Background}

\subsection{Improving moral decision-making (MDM)}

We adopt the definition given by Garrigan and colleagues \cite{garrigan2018moral} that moral decision-making (MDM) is ``any decision, including judgements, evaluations, and response choices, made within the `moral domain''' with the moral domain consisting of decisions concerning issues like harm and fairness. The authors show how theories for the development of morality in humans fall into three main categories: cognitive, affective (or emotional), and social \cite{garrigan2018moral}. Cognitive theories take on a neuroscience-based approach (i.e. which parts of the brain are activated in response to moral stimuli), affective theories are based primarily in developmental psychology, and social theories reflect how moral psychology and behaviours changes from an individual to a population \cite{garrigan2018moral}. We recap the authors' original summary of the various theories to illustrate how they have evolved in complexity. Theories from developmental psychology look at how moral reasoning develops, firstly in childhood \cite{piaget1932moral}, then beyond into adolescence and adulthood \cite{kohlberg1976moral}, and now incorporate a spectrum based in cognitive theories like perspective taking and scripts \cite{rest1999postconventional} as well as attention and working memory \cite{gibbs2013moral}. The most well-known affective theory - Haidt's social intuitionist theory \cite{haidt2001emotional, haidt2008social} - investigates how quicker emotional decisions are rationalised post-hoc into ``moral'' decisions (akin to System 1 of Kahneman's reflexive System 1 and more calculating System 2 ways of thinking \cite{kahneman2011thinking}). Other affective theories like dual-process theory have incorporated neuroscience aspects by focusing on activity levels of parts of the brain with real-time MDM \cite{greene2001fmri, greene2002and}. 

MDM necessitates navigating trade-offs between different interests, such as those of individuals, groups, or society as a whole \cite{gauthier1987morals, vitell1993effects, cribb2011shared}. This makes MDM emotionally challenging since it involves choices that have significant consequences for oneself or others with positive and negative effects often getting amplified with scale \cite{berman2020moral}. The inclusion of AI models in emotional decisions may initially seem off-putting \cite{helberger2020fairest}. But the emotional challenge of making important decisions is the exact reason that we want non-emotional agents involved - so they can minimise human inconsistency \cite{asch1956studies} and provide fairer outcomes \cite{lipton2018mythos, rudin2019stop}. We are careful here to avoid referencing ``automation'' of human decisions \cite{birhane2021impossibility}. Real-world decisions have multiple stages which cannot currently, and likely never will, be fully automated by AI models. Instead, we look to improve moral decision making by integrating AI into standard human decision-making processes in stages which are prone to human error or with data that are beyond our cognitive capacities \cite{suresh2021beyond, cai2019hello, feng2016synthesis, araujo2020ai}. So, how do we integrate moral psychology theories into our AI models? Moral philosophies such as Deontology allow us to formalise aspects of MDM but the multi-factorial development of morality in humans is hard to represent as just one moral philosophy \cite{upton2009virtue}. Context-specific models may be enabled by singular theories such as Virtue Ethics for general clinical settings \cite{hindocha2022moral} or Bentham's Felicific Calculus for end-of-life situations \cite{post2022breaking} but more flexible constructions are also possible. These flexible constructions help generalise MDM systems to unseen and novel moral situations \cite{jin2022make}. We outline the overarching study of these models, called artifical moral agents, in the next section.

\subsection{Different constructions of artificial moral agents (AMAs)}

The study of artificial moral agents (AMAs) is an interdisciplinary field between computer science, ethics, and philosophy. As such, we first clarify terminology. The terms “model” and “agent” both refer to AI systems, and we use agent to emphasise that the model has a degree of autonomy. ``Morals'' concern actions of virtue and ``morality'' reflects that these behaviours are practised habitually to become things we accept internally and externally as rules or principles \cite{cervantes2020artificial} - more concisely, morals ``regulate selfishness and make social life possible'' \cite{haidt2008morality}. ``Ethics'' is a broader term than morality which can be defined as a ``rational reflection on moral behaviours'' \cite{cervantes2020artificial} and better emphasises contextual differences for moral behaviours \cite{mattingly2018anthropology}. The words are closely linked and for our purposes can be used interchangeably, but we refer solely to morality going forward since this is the terminology of AMAs. Thus, an AMA is a program that can act or make decisions in a “moral” way, with a degree of autonomy \cite{cervantes2020artificial}. The autonomy of an AMA is the extent to which a human can interact with the agent to change one of its decisions \cite{mostafa2019adjustable}. There are three categories of AMAs distinguished by the level of moral consideration built into them and that they can act on: implicit, explicit, and full \cite{moor2006nature}. Implicit AMAs cannot distinguish good from bad behaviour but are constructed to enable moral behaviour, explicit agents use inbuilt ethical rules (e.g. from logical formalisms or algorithmic constraints), while full ethical agents, like humans, possess aspects of consciousness like desires, intentions, and free will. We only consider AMAs within the first two levels to ignore questions related to fair treatment of potentially sentient artificial agents, limiting the scope of interpretability requirements to human (and not machine) safety. The breadth of these categories makes it more challenging to analyse how their differences manifest in real-world agents so we turn our attention to more granular parameters of AMA construction: the moral paradigm, the scale, and the purpose of the agent.

We group moral philosophies and moral psychologies under the name moral paradigm or framework which tell us how morality is instilled into the agent. There are three broad moral paradigms we consider: top-down (TD), bottom-up (BU), or hybrid. TD approaches start from a set of principles or a moral framework (e.g. Utilitarianism), BU approaches have no moral framework and instead aim to learn morality from the environment, and hybrid approaches combine aspects of the two \cite{allen2005artificial}. Agents, like standard AI models, can be constructed at different scales which produce different performance capabilities \cite{provost1996scaling} - we consider standard individual agents, high capacity individual agents (vertical scaling), and multi-agent systems (horizontal scaling) \cite{ali2019survey}. For simplicity in horizontally-scaled systems, we assume all agents are cooperative and that there are no unpredictable agent-agent interaction effects \cite{cervantes2020artificial}. The purpose of the AMA is the task that it is designed to do and can be split into uni-purpose, multi-purpose, and general-purpose. The distinction between multi-purpose and general-purpose gets blurred as agents become more capable at multiple tasks and so, to avoid case-by-case analysis of different purposes, we focus on the distinction between uni-purpose and general-purpose.

\subsection{Interpretability ``levels'' and their importance for MDM}

There is no agreed-upon definition for interpretability but it can be viewed generally as a domain-specific quality for understanding or trusting our agent \cite{rudin2019stop, lipton2018mythos}. Two seminal perspectives from the explainability/interpretability literature pose a dichotomy where ``explainability'' is using a black box model and then explaining it with a secondary post-hoc model, and ``interpretability'' is not using a black box, instead using a model that explains itself (a.k.a. a white box or transparent model) \cite{rudin2019stop, lipton2018mythos}. \citet{lipton2018mythos} adds further detail based on three different paradigms of white box modelling: algorithmic transparency, decomposability, and simulatability. Algorithmic transparency amounts to a formal understanding of the agent's learning process \cite{lipton2018mythos}, for example, better characterisation of the loss surface \cite{garipov2018loss, fort2020deep} or providing internal convergence properties \cite{jacot2018neural}. Decomposability corresponds to transparency at the level of model parameters while simulatability is transparency across the whole model \cite{lipton2018mythos}. Within the context of MDM, decomposability corresponds to having an intuitive and step-by-step explanation for each major agent decision \cite{nashed2023fairness}. We believe that complete decomposability (i.e. intuitive explanations for all model parameters and output decisions) subsumes simulatability and so do not consider simulatability further. For clarity, we use “transparency” when referring to forms of white box agents, “post-hoc explainability” for explanations of black box agents, and “interpretability” when referring to both of these concepts together.

Interpretability alone is not necessarily useful for our domain of MDM but becomes so when directed towards a specific goal \cite{krishnan2020against}. Watson lays out three challenges that interpretability faces \cite{watson2022conceptual}: clarity of what the explanation corresponds to (e.g. the model's outputs, the data generating process, different sub-objectives), error rates and consistency rates for explanations, and little consideration of the fact that explanations can change over time. Given these challenges are important for all interpretable systems, we take the aim of interpretability in MDM as providing an understanding of agent decision-making processes for appropriate error correction, error prevention, and agent behaviour optimisation. More simply, interpretability is useful here as a debugging tool for different stakeholders involved in a moral decision \cite{suresh2021beyond}. While there are clearer situations where interpretability is not needed, such as when an agent is not involved in a high-stakes decision \cite{rudin2019stop} or does not have a significant impact on society \cite{molnar2020interpretable}, we assume all decisions requiring moral consideration as potentially high-stakes and enabled by reliable human-agent collaboration \cite{feng2016synthesis, dietvorst2015algorithm, araujo2020ai}, and thus in need of some form of interpretability. This lends itself to our characterisation of interpretability as a spectrum more than a binary requirement. 

While the different types of interpretability do not fall neatly into a hierarchy of explanation complexity \cite{lipton2018mythos}, this becomes easier when each type is viewed as a debugging tool. Certain types of interpretability are more challenging to program into an agent and different types are required depending on what the AMA does and the number of stakeholders involved. A loose interpretability hierarchy in terms of increasing agent construction difficulty, which we phrase as ``levels'', is as follows: black box models - post-hoc explanations of black box models - algorithmic transparency - decomposability. 

Below, we address exactly these different facets of interpretability, starting with asking ourselves about whether black box AMAs allow for trust and continuing with whether transparency is the key to understanding all AMAs.

\section{Does lack of transparency in AMAs prevent trust?} \label{can}

In this section, we discuss the reliability of moral reasoning in AMAs without transparency, commonly seen as black box AMAs. Thus, we will use ``black box" as a generic term for such systems. Since the internal agent reasoning is unavailable to us directly, we require a level of faith in our agent, which corresponds to framing \textbf{interpretability as trust} for MDM \cite{lipton2018mythos}. Trust can take on a range of meanings: confidence in the agent to make the correct decision, the consistency of the agent's decisions in certain situations, or whether the agent makes decisions that are right or wrong in a human-like manner. We discuss whether AMAs can learn moral principles, and then if so, whether these principles are appropriate. We conclude that we can trust AMAs without transparency if they adopt the benefits of both BU and TD agents.

\subsection{Can we tell if black box AMAs have learned \textit{any} moral principles?}
\label{can-they}

We define the ``environment'' of an AMA as the potential hypothesis space spanned by the data and the AMA's learning process, which consists of its internal model and training regime. BU AMAs are predicated on the existence of functional morality \cite{johansson2010functional} - that agents are able to learn morals from their environment. With a black box BU AMA, how can we be sure that our agent has learned some form of morality? \citet{allen2000prolegomena} defined a Moral Turing Test which states that “if two systems are input-output equivalent, they have the same moral status”, with the subsequent debate following that of the Chinese Room Argument \cite{searle1980minds}. Effectively, any machine capable of memorising a sufficiently diverse and framework-like or human-like set of input-output moral relationships would pass the test, even though we would not be able to determine if it is intrinsically “moral”.

There is no universal definition for morality in humans beyond notions of obligation \cite{skorupski1993definition}. However, psychological theories for the development of morality within humans \cite{hardy2011moral, garrigan2018moral} and the ``teachability'' of moral values \cite{prior1990virtue, straughan1988can} point to the development of morality as a process \cite{garrigan2018moral}, if an imprecise one. It is from the process of learning via experience in the world that we as humans feel the obligation to be a moral agent \cite{skorupski1993definition}. The same argument has been made for AI models \cite{badea2022morality} through development of value-based agents which learn human values, a process which has been successfully implemented in the multi-valued action reasoning system \cite{badea2022have}. The lack of a universal definition means that memorisation could suffice as a type of morality, particularly if we consider memorisation a form of learning \cite{hoque2018memorization}. However, it is insufficient for trusting our agent because there are no guarantees that the agent will generalise outside the training set. This is an issue that can be explained more clearly by the difference between form and meaning as expressed initially by Badea and Artus \cite{badea2022morality} and subsequently formalised by Bender and colleagues \cite{bender2020climbing}. They state that if an agent is trained only on form (e.g. pixels, words, etc.) without any input for communicative intent behind the form (i.e. context-dependent meaning, which exists at different scopes of worldly experience \cite{bisk2020experience}), it cannot truly intuit any meaning of morality, and certainly no definition that extends to novel contexts \cite{garrigan2018moral, jin2022make, shen2021towards}.
This comes down to the ``The Interpretation Problem'' \cite{badea2022morality}, which refers to the issue of endless potential interpretations for any symbolic representation given to an AMA. This makes it impossible to guarantee a fully accurate transmission of meaning regardless of the medium chosen. 

The question then arises: would sufficient memorisation of triples with a structure of (input, output, communicative intent) suffice for ``learning morality''? The ever-changing nature of communicative intent across cultures and over time requires a potentially infinite and unobtainable set of such data \cite{bisk2020experience} which prevents learning a cohesive set of moral principles \cite{bender2020climbing} unless the context is clearly defined a priori and an acceptable level of error with an ``action limit'' defined \cite{sculley2014technical}. To avoid the difficulty of defining a complete and ethical training dataset for an imprecise objective \cite{dignum2018ethics, hendrycks2021aligning, jin2022make, sap2020social}, we can instead use black box TD AMAs to approach the problem from a different angle: ensure the agent has a set of pre-defined moral principles rather than relying on the data and where it has come from. TD AMAs use a specific moral framework (or set of frameworks) which allow us to compare the TD agent's input-output moral relationships with the most likely output from that same moral framework. While the TD construction is more rigid than the BU construction, it enables greater trust in the AMA's learned principles because we have a form of ground truth that is less variable than individual agent comparisons permitted by the Moral Turing Test for BU constructions. Hybrid settings require additional domain knowledge but are even better since they can find the right ``compromise between being too flexible and too strict'' \cite{morley2021ethics}. From this, we deduce that morality can be learned by agents once it has some initial framework for a given domain, and further capacity can be enabled by data with communicative intents.

\subsection{Can we guarantee that black box AMAs have learned \textit{appropriate} moral principles?} \label{appr}

Let us assume that our AMA is able to learn moral principles from its environment. We are now presented with a different problem: how can we be sure that this environment reflects our desired human values and that the agent is learning them? Current inequalities in our world have been shaped (and are still influenced) by cultural remnants of historically unequal power dynamics \cite{tilly2005historical, kenfack2021impact}. An overt example is historical medical exploitation of underrepresented communities which has led to a lack of diversity in large-scale genomic data, and been a major obstacle to generalisable genomic insights across populations \cite{fatumo2022roadmap}. Similarly damaging, but more subtle, is the inadequate treatment of sensitive variables (e.g. age, sex, race, etc.) which can lead to models shortcutting to high predictive accuracy based on harmful stereotypes \cite{geirhos2020shortcut, rudin2019stop} - textbook examples include racist explanations in criminal recidivism prediction \cite{chouldechova2017fair} and proxy variables that consolidate racial disparities in population health models for medical support prioritisation \cite{obermeyer2019dissecting}. Beyond error correction, models can also serve as ways to improve existing disparities - for example, enabling smoother socioeconomic mobility via smarter intergenerational wealth allocation \cite{heidari2021allocating}. Given that these inequalities persist in both data collection and the data itself \cite{ricaurte2019data}, potentially in implicit ways, any AMA reliant on its environment has the potential to propagate or even amplify these inequalities. As computer scientists, we have the opportunity to build algorithmic mechanisms into our AMAs to counteract and help remedy these types of bias \cite{mohamed2020decolonial}. If we take this pro-active approach to equality, it becomes important to understand how biases exist in our environment, how these get encoded in our data, and how the AMA can use them inappropriately in its reasoning or explanation mechanisms \cite{schwartz2022towards}. This is more important in BU systems due to their higher capacity for learning morality from the environment \cite{cervantes2020artificial}. 

For systemic inequalities that affect marginalised communities, minimising predictive disparities over different demographics is a proxy for our AMA ``learning'' appropriate moral principles. This line of work has been explored extensively in the fairness and sequential decision-making literature \cite{nashed2023fairness} and we briefly review important instances. Ensemble models have proven effective, with different weighting schemes used per classifier \cite{kenfack2021impact, gohar2022towards} and for ``unfairly classified'' samples \cite{bhaskaruni2019improving}. Coston and colleagues \cite{coston2021characterizing} found that characterising properties over the Rashomon set - an ensemble of models that all perform highly which, in this case, is the set of most fair models - gave them algorithmic bounds for the range of disparities with applications to recidivism risk prediction and consumer lending. Beyond outcome evaluation, \citet{dai2022fairness} laid out metrics for evaluating post-hoc explanations: fidelity, stability, and consistency, aligning with the main conceptual challenges for IML \cite{watson2022conceptual}. They also evaluated the practical explanation quality of sparsity, which is a proxy for how ``understandable'' an explanation is, with higher sparsity allowing for fewer features and thus easier understanding. These ideas of human-intuitive explanations are expanded on in Section \ref{decomp}. Well-aligned explanations are useful because they can reduce overreliance on AI systems and make human-AI interaction more coordinated \cite{vasconcelos2023explanations}.

Taking a more engineering-based approach, \citet{shaw2018towards} defined meta-qualities for desired moral standards to guide BU AMAs under very constrained applications. Accepting that uncovering the reasoning of individual agents is challenging, the authors analysed multi-agent systems via post-hoc explainability to derive bounds for moral behaviour. These multi-agent behaviours can also be viewed from a TD lens via consideration of TD agents and stakeholders in a complete sociotechnical system \cite{murukannaiah2020new} or collective actions of TD agent systems called moral communities \cite{nashed2021ethically}. The bounds in these cases are based on average population behaviour and act like probabilistic alternatives to the Moral Turing Test. Stronger fairness tests, such as those based on localised program execution paths \cite{aggarwal2019black}, would be needed for individual instances of discrimination. A more intuitive way to view these ``moral behaviours'' is as highly likely probabilistic constraints - similar to the way we would view the chance of a bridge breaking under stress - they should hold under all reasonable perturbations within a given context. 

The limitation of TD AMAs is that it is challenging to select the appropriate moral paradigm for a situation, made more difficult by the fact that there are several potentially appropriate moral decisions (i.e. input-output relation cardinality of moral decisions is one-to-many) based on varying sequential decision trajectories. Reinforcement learning (RL) has become the de facto toolkit for sequential decision-making since it can comprehensively explore a given decision space \cite{nashed2023fairness} - in moral agent terms, this amounts to BU flexibility within wide-ranging but well-defined TD constraints, or the hybrid moral paradigm. For robustness to this decision trajectory variation, \citet{svegliato2021ethically} developed an AMA based on RL that uses all previous decision states to make its final decision. Notably, this AMA also circumvented the issue of imprecise objective functions by decoupling the moral compliance objectives from the task objectives. This decoupling was also recommended as one of the main ways to circumvent the Interpretation Problem in \cite{badea2022morality} and mirrors their distinction between moral mistakes and amoral mistakes. This avoids issues of ambiguous fidelity with respect to explanations \cite{watson2022conceptual}. In a less constrained decision-space than the setting in \citet{svegliato2021ethically}, trajectories can get more unwieldy. Subsequent work by \citet{srivastava2023planning} applies additional constraints to the agent (instead of its environment) which can analyse negative side effects of prototype decision sequences with human input and then replan an appropriate sequence which minimises these side effects. RL can also modify the environment to make it more appropriate. Using a reward that includes both the performance and moral compliance objectives, \citet{rodriguez2022instilling} refine the convex hull of a Markov decision process (the stochastic process underlying RL) to get moral bounds on the decision space, which also mirrors the second suggestion for circumventing the Interpretation Problem.

Having begun this section with a discussion of fairness requirements, we covered how they can be enabled through metrics on the black box outputs, surveyed moral explanations for black boxes, and finished by reviewing high capacity black boxes for sequential decision-making. Regardless of the chosen interpretable black box paradigm, AMAs can be trusted without transparency if their level of capacity is well-tuned to their purpose. However, steps should be taken to enable debugging where applicable. Fairness and moral compliance objectives should be distinct from performance objectives \cite{rossi2019building, svegliato2020integrated}, and ideally, post-hoc explanations should be used to facilitate easier debugging in case of faulty agents. As a general rule, explanations or decision trajectories should be stress-tested in different scenarios and their consistency analysed \cite{watson2022conceptual}. Thus, for additional safety, we recommend that the MLI for trustworthy black box models be consistent explanations or decision trajectories over important subgroups of the populations in the dataset. 

\section{Does transparency help us understand AMAs?}

In this section, we frame interpretability as transparency for internal model reasoning, looking at two forms of transparency: \textbf{algorithmic transparency} and \textbf{decomposability} \cite{lipton2018mythos}. With that in mind, we show below that the utility of transparency varies in magnitude and is context-dependent. We propose algorithmic transparency as single-agent rules composed into multi-agent rules. Furthermore, we argue that the importance of transparency rises with the causal power of the agent, and depends on the relevant stakeholders. 

%TODO: check if reneira is needed or not
In real-world settings, AMAs require higher capacity to interact and respond to their environment \cite{badea2022morality}. For this, we assume that the moral paradigm (BU/TD/hybrid) of the agent is sufficiently flexible to allow adaptation to new environments. With that assumption, the most important AMA construction parameters for analysing deployed AMAs become the scale and purpose of the agent. We note these are both somewhat nebulous terms that incorporate aspects outside AMA construction: `scale' encompasses the number of model parameters, the capacity to act in the world, and the number of other agents which interact with it (for instance its users) and `purpose' can be defined a priori by developers via its task objectives and moral paradigm but is ultimately at the hands of the user. Additionally, the purpose of an AMA is dynamic and can change with regards to performance capabilities achieved at scale. For added clarity when considering both construction parameters and transparency terms in the following sections, we centre the discussion of algorithmic transparency on AMA construction since specifics of the explanation form are less important, and we centre decomposability on AMA users since the explanation form is paramount to its utility.

\subsection{The utility of algorithmic transparency depends on AMA construction}
\label{algo-transparency}

The Artus-Badea law states that an increase in scale gives an AMA more causal power and so more exposure to risk \cite{badea2022morality}. Thus, if an AMA does not have a significant effect on the world around it, there is less of a safety requirement for transparency to understand the agent's reasoning \cite{molnar2020interpretable}. Therefore, in such trivial cases, one could get away with not implementing any explicit transparency. But moral ``significance'' is not always obvious because of collective agent behaviour from horizontally-scaled systems. For example, say you design an agent for your own use to analyse the sentiment of current news (whether the news is positive or negative) so you can prioritise more positive news stories. This only involves you and is thus unlikely to have a direct and significant impact on society, and would be a prime candidate for potential safety without transparency. However, if you decide to scale the model up or make it into a commercial product so that other people use the same agent, then moral questions arise because its effects on society are compounded and individuals might experience different uni-agent effects. For example, we have the development of echo chambers within social media websites, the reinforcement effect this has on sub-populations (perpetuating their existing opinions), and then the combined polarising effect on the entire population (the radicalisation of opposite sides) \cite{cinelli2021echo}. 

Now onto our proposed MLI for this case. As discussed in Section \ref{appr}, multi-agent system behaviour used as post-hoc explainability can help give us guarantees on general agent morality \textit{for specific tasks}. However, without some level of individual agent transparency, we have no guarantees on agent subgroup behaviour below a certain (unknown) subgroup size, and consequently a deeper understanding of the overall population becomes intractable. Internal ensembling over outputs is a practical way to get probabilistic limits to behaviour \cite{tekin2016adaptive} for black box singular agents and to mitigate this population-subgroup mismatch. We have discussed algorithmic fairness over ensembles in function space \cite{coston2021characterizing} (i.e. potentially vastly different models) but \citet{barnett2023active} propose a simpler alternative within a reinforcement learning framework that ensembles over two models predicting opposite things. They compare the output of rational and irrational teacher models to quantify the difference between them and thus produce a better ``rationality direction'' for future decision trajectories. Accordingly, we recommend the MLI for horizontally-scaled AMAs to be a nested combination of black box interpretability. Thus our proposal is thinking of algorithmic transparency as logical rules \cite{mermet2016formal, rossi2019building} (or as probabilistic limits) \cite{shaw2018towards, rossi2019building} intended for uni-purpose individual agents and composed into multi-agent rules (or limits). In other words, this means bounding multi-agent systems by bounding uni-agent systems with algorithmic transparency. As the agents themselves scale vertically (i.e. their number of parameters increases), the range of behaviours each agent can perform increases and, when combined in multi-agent systems, results in complex collective behaviour \cite{kelly2009out}. To limit the complexity of our discussion, we do not talk about this combined vertical and horizontal scaling case further, instead moving to focusing on vertically-scaled agents.

Vertical scaling of models is performed through increasing three parameters: training data size (with the assumption that quality stays the same), compute power, and model capacity \cite{kaplan2020scaling}. Although algorithmic transparency has not been studied for vertically scaled AMAs directly \cite{cervantes2020artificial}, we can use large language models (LLMs) as an approximation given their ongoing integration into MDM settings like medicine \cite{ayers2023comparing, lee2023benefits}. LLMs such as GPT-3 have demonstrated linear scaling laws for prediction, that is: as the three parameters increase, the performance of the model increases linearly \cite{kaplan2020scaling}. However, \citet{ganguli2022predictability} showed that although these scaling laws hold at a general level across multiple tasks, performance on specific tasks can change abruptly at arbitrary scaling points of the three parameters, raising questions on what the models are actually learning. By reverse engineering neural networks, we are beginning to mathematically understand these flows of information \cite{elhage2021mathematical} and discontinuous jumps to qualitatively better performance \cite{nanda2023progress} during the learning process. However, these results are currently limited to toy models (one layer multi-layer perceptrons and attention modules) of much smaller size than those in deployment and there is thus uncertainty about their generalisation to real-world models with multiple components \cite{nanda2023progress}. Regardless, the rapid uptake and potential of these models necessitates guarantees on MDM to lessen harmful effects on end-users \cite{weidinger2021ethical}. Where these guarantees are not currently expressible as a proof or formula, we propose expressing them as probabilistic guarantees or qualitative explanations that reveal opaque agent reasoning. As such, we continue the discussion of vertically-scaled agents via decomposability in large language models in the following section.

\subsection{The utility of decomposability depends on the stakeholder}
\label{decomp}

What makes an explanation of an agent's decision-making intuitive? As has been a common theme throughout this paper, there is no single answer since intuitive explanations are dependent on the type of stakeholders and their level of interaction with the agent. Suresh et al. \cite{suresh2021beyond} ascribe two essential facets to stakeholders in interpretability: their level of expertise (knowledge within a context), and their goals in the long- and short-term. While long-term goals (model understanding and trust) are common across all stakeholders, short-term goals can differ. Given our focus on MDM and user accessibility, we are focused on the (O1), (O7), and (T1-4) short-term goals (as defined in \cite{suresh2021beyond}) which revolve around model debugging, improvement, and feature importance \cite{suresh2021beyond, miller2022stakeholder}. 

As part of pursuing these goals, we segment our stakeholders into two functional categories - developers and end users - for a clear and concrete analysis of stakeholder concerns in interpretable MDM. Intuitive explanations of agent reasoning steps are important for developers and users alike so that both can easily modify the AMA to rectify and prevent unintended behaviour (i.e. error reduction) while also fine-tuning the agent for more desirable behaviour (i.e. agent optimisation). To better align the actions of high-capacity AMAs to human values, we have touched on algorithmic mechanisms of counteracting bias and designing modular objective functions (moral and performance), but these interventions are restricted to developers. To make alignment accessible to the end users, we can instead express these algorithmic changes directly through text or other intuitive modality forms for humans (e.g. audio, video) \cite{chen2022holistic}. One advantage is that this can help level the commercial and regulatory playing field between developers and users. \citet{rudin2019stop} describes how the black box nature of models allows companies and their developers to get away with some faulty individual predictions if their average behaviour is sufficient, which the above proposal mitigates. We propose integrating decomposability into AMAs, which would mean that explanations would be presented in a way that allows users to understand the AMA's reasoning, giving them more control and customisation over the agent, and decreasing the chance that developers would be able to exploit them through asymmetric information \cite{molnar2020interpretable}. 

In the same vein, we believe that the success of publicly-deployed LLMs like ChatGPT is largely due to their use of text as an interface for users. The input text for LLMs is called a ``prompt'' - the ease of prompt design and its interpretations as both probabilistic and textual inputs have inspired new ways of formalising LLMs at multiple interpretability scales \cite{liu2023pre, dohan2022language}. A key paper from this line of research is Chain-of-Thought (CoT) prompting \cite{wei2022chain} which involves setting up the prompt with sequential reasoning steps (a ``rationale chain'') to lead the LLM to give its response using a similar rationale chain. Extensions to CoT have focused on automating rationale chain generation via pre-defined prompt phrases to generate multiple rationale chains \cite{huang2022large} and smart pruning of less likely rationale chains \cite{shum2023automatic}. However, CoT reasoning is an emergent phenomenon of large models ($>$100 billion parameters). For medium-sized LLMs (10-100 billion parameters), Anthropic have found that they can learn moral concepts related to harm like stereotyping and bias when given clear instructions \cite{ganguli2023capacity}. 
Going one step further, \citet{jin2022make} developed a CoT extension which determines when it is appropriate to break moral rules with a provided rationale, giving textual explanations for both developers and users. Currently, this work is limited to three lower-stakes situations within the cultural context of the USA but the experiments are initial proof that medium-sized (and larger) LLMs are capable of learning and reasoning about moral obligations \cite{hendrycks2021aligning, jiang2021delphi, forbes2020social}. This suggests that larger LLMs can also be fine-tuned for specific moral consideration by all stakeholders with appropriate datasets. 

The arguments above show once again the value of our proposal of using decomposability to enhance stakeholder accessibility and enable them to do more. Having explained the importance of decomposability, we also describe its fundamental limitation: \textit{oversimplification}. Humans are complex adaptive systems that exist within the dynamics of societal interaction \cite{birhane2021impossibility}. This complexity means digitisation (or conversion into ``form'' \cite{bender2020climbing}) of context-dependent human concepts like trust, understanding, or morality is just an approximation of the real thing \cite{birhane2021impossibility}. Explaining these digitised concepts, whether internally during processing or post-hoc, will only also be further approximations because explanations are simplified representations of the original model \cite{rudin2019stop, sap2020social}. Inputs to the agent need to contain causal information for outputs, otherwise explanations of morality cannot translate from humans to machines and back without a loss of information, making them brittle and particularly harmful to long-term trust \cite{papagni2020interpretable}. This may seem like a pessimistic view of interpretability, but it draws attention to the fact that our internal representations of moral principles and their subsequent preservation through AMA processing are the keys to ensuring useful explanations. Referring back to our discussion of form and meaning in Section \ref{can-they}, useful model decomposition will require adaptive multi-modal agents grounded in experience of the world \cite{chandu2021grounding, bisk2020experience}.

\section{Conclusion}

On the one hand, as we argued earlier, artificial moral agents (AMAs) can be created without interpretability and be given sufficient trust for moral reasoning in some narrow and well-defined tasks. The Interpretation Problem \cite{badea2022morality} means that we would still never get perfect guarantees about moral behaviour but we can get around this by building value-based agents that can be tested for trustworthiness \cite{badea2022have}. However, relying on ``input-output'' tests, like the Moral Turing Test, limits this trust because they do not evaluate for ``intrinsic'' morality or the possibility of several acceptable moral outputs to the same inputs. To improve this and make individual agents more reliable, we can use principles based on collective agent behaviour to guide them and address this issue of trust in black box bottom-up agents with opaque internal mechanisms \cite{shaw2018towards}. Additionally, top-down black box AMAs allow us to define prior moral constraints and carefully construct objective functions in AMAs so their moral reasoning is more consistent and predictable \cite{svegliato2021ethically}. 

On the other hand, for general-purpose AMAs, we need stronger levels of interpretability requirements. While we have seen large-scale studies that show it is possible to reliably obtain good general AMA behaviour, a poor understanding of the inner mechanics of these agents has still resulted in abrupt scaling issues and unintended individual agent behaviour \cite{ganguli2022predictability, weidinger2021ethical}. For optimal levels of trust in and between agent-developer-user systems, explanations at different levels of abstraction, while imperfect \cite{birhane2021impossibility, rudin2019stop}, are imperative to help these AMAs reach safe deployment. Importantly, for future work in this area, better quantification of moral compliance can decisively aid the understanding of interpretability requirements in different contexts, while neurosymbolic methods can help with constructing top-down and hybrid AMAs \cite{de2021aligning, roy2022proknow} and causality across the three rungs of Pearl's ladder \cite{pearl2009causality} with more comprehensive understanding of moral decision-making \cite{kusner2017counterfactual, mhasawade2021causal, nashed2022unifying, geiger2022inducing}.

In conclusion, while trustworthy AMAs can be created without any level of transparency, they make rigorous assessment and risk mitigation of AMAs much more challenging. For the moral paradigm of an AMA, we recommend top-down or hybrid agents, advocating against the use of bottom-up agents due to their higher risk of learning improper moral principles when deployed in novel environments. For AMAs with variable purposes, we believe algorithmic behavioural guarantees are the MLI for \textit{uni-purpose AMAs}, with additional during-processing explanations, or task-specific decomposability, being the MLI for \textit{general-purpose AMAs}. Additionally, for both of these purposes, moral compliance objectives should be as distinct from performance objectives as possible for easier ``quantification'' \cite{watson2022conceptual, badea2022have}. For scale, both horizontally- and vertically-scaled systems require strong algorithmic behavioural guarantees, and for those with multiple stakeholders, intuitive explanations in both algorithmic and textual forms, or stakeholder-specific decomposability. The general rule that we propose builds upon the Artus-Badea law \cite{badea2022morality}, and is one of common sense: the higher capacity an AMA is, the more potential it has for a wider user base, the more safety and interactivity mechanisms are needed, and so a ``higher'' Minimum Level of Interpretability is required.

\section{Acknowledgements}

A.V. is supported by a UK Research and Innovation (UKRI) Centre for Doctoral Training in AI for Healthcare PhD studentship (EPSRC, EP/S023283/1).


\bibliography{sn-bibliography}% common bib file

\end{document}